# General-Purpose MCMC Inference over Relational Structures

**Brian Milch**
Computer Science Division
University of California
Berkeley, CA 94720-1776
milch@cs.berkeley.edu

**Stuart Russell**
Computer Science Division
University of California
Berkeley, CA 94720-1776
russell@cs.berkeley.edu

## Abstract

Tasks such as record linkage and multi-target tracking, which involve reconstructing the set of objects that underlie some observed data, are particularly challenging for probabilistic inference. Recent work has achieved efficient and accurate inference on such problems using Markov chain Monte Carlo (MCMC) techniques with customized proposal distributions. Currently, implementing such a system requires coding MCMC state representations and acceptance probability calculations that are specific to a particular application. An alternative approach, which we pursue in this paper, is to use a general-purpose probabilistic modeling language (such as BLOG) and a generic Metropolis-Hastings MCMC algorithm that supports user-supplied proposal distributions. Our algorithm gains flexibility by using MCMC states that are only partial descriptions of possible worlds; we provide conditions under which MCMC over partial worlds yields correct answers to queries. We also show how to use a context-specific Bayes net to identify the factors in the acceptance probability that need to be computed for a given proposed move. Experimental results on a citation matching task show that our general-purpose MCMC engine compares favorably with an application-specific system.

## 1 INTRODUCTION

Many probabilistic reasoning problems involve reconstructing the set of (possibly related) real-world objects that underlie some observed data: for instance, the publications and authors that are mentioned in a set of bibliographic citations, or the aircraft that underlie a set of observed radar blips. Because the number of possible ways to map a set of observations to underlying objects is huge, these problems are extremely challenging for standard proba-bilistic inference algorithms. Specialized algorithms have been developed in the fields of record linkage [Fellegi and Sunter, 1969] and data association [Bar-Shalom and Fortmann, 1988] that yield approximate solutions for particular classes of models.

Recently, Pasula *et al.* [2003] and Oh *et al.* [2004] have achieved state-of-the-art results on these problems using Markov chain Monte Carlo (MCMC) methods. Given a probability distribution $p$ on an outcome space $\Omega$, an MCMC algorithm approximates the probability of a query event $Q$ given an evidence event $E$ by generating a sequence of samples $s_1, s_2, \ldots, s_N$ from a Markov chain over $\Omega$. This Markov chain is chosen so that it only visits outcomes consistent with $E$, and its stationary distribution is proportional to $p(s)$. The desired probability $p(Q|E)$ is then approximated as the fraction of $s_1, \ldots, s_N$ that are in $Q$. If the Markov chain is ergodic, then this approximation converges to the correct posterior probability as $N \to \infty$.

Pasula *et al.* [2003] and Oh *et al.* [2004] use the Metropolis-Hastings (M-H) algorithm [Metropolis *et al.*, 1953; Hastings, 1970] to construct a Markov chain whose stationary distribution is proportional to $p(s)$. In the M-H algorithm, when the current state of the chain is $s_n$, a new state is sampled from a *proposal distribution* $q(s'|s_n)$. The *acceptance probability* for this proposed move is:

$$\alpha(s_n, s') \triangleq \max\left(1, \frac{p(s')q(s_n|s')}{p(s_n)q(s'|s_n)}\right) \quad (1)$$

With probability $\alpha(s_n, s')$, the move is accepted and $s_{n+1}$ is set to $s'$; otherwise, $s_{n+1} = s_n$. The proposal distribution can incorporate domain-specific heuristics that help the chain to move quickly among reasonably probable hypotheses. As long as this Markov chain is ergodic, the stationary distribution will be proportional to $p(s)$ regardless of the proposal distribution.

The recent successes of the M-H algorithm suggest that it could be applied usefully to other relational inference tasks, such as resolving coreference among noun phrases in text. However, to recycle a comment that Gilks *et al.* made about Gibbs sampling in 1994, "Until now all existing implemen-tations have involved writing one-off computer code in low or intermediate level languages such as C or Fortran." Although perhaps Matlab has replaced Fortran, this is still true about M-H implementations today: the data structures that represent MCMC states and the algorithms that compute acceptance probabilities are application-specific. Tackling a new application — or even adding a variable to an existing model — requires rewriting portions of the state representation and acceptance probability code, in addition to modifying the proposal distribution.

We would prefer to have a general approximate inference system that computed answers to queries based on a user-supplied probability model and proposal distribution. This paper presents a prototype for such a system. To represent the model, we use Bayesian logic (BLOG) [Milch et al., 2005a], a language that allows us to express uncertainty about the number of latent objects and the relations among objects and observations. Many of our results would also be applicable to other probabilistic modeling languages [Pfeffer, 2001; Jaeger, 2001; Laskey and da Costa, 2005; Richardson and Domingos, 2006; Mjolsness, 2005]. Ideally, we would also like to have a declarative language for specifying proposal distributions. However, at this point we are only beginning to understand what constructs and features such a language would need to have. Thus, we assume that proposal distributions are specified procedurally, as Java classes that implement a certain interface.

This paper focuses on the interface between the proposal distribution and the main M-H engine — specifically, the representation of MCMC states — and on how the engine can efficiently compute acceptance probabilities for arbitrary proposals. We begin in Sec. 2 by describing the probability model and M-H implementation used by [Pasula et al., 2003] to resolve coreference among citations. We then briefly review the BLOG language of [Milch et al., 2005a]. Sec. 3 introduces the architecture of our generic M-H system. Our main contributions begin in Sec. 4, which discusses the state space of our M-H algorithm. The key point here is that it is impractical to use MCMC states that correspond to single possible worlds; instead, states are represented with *partial* descriptions that denote whole sets of worlds. We provide conditions under which MCMC over sets of worlds yields asymptotically correct answers to queries. Taking advantage of this theorem, we use state descriptions that are partial in two ways: they do not instantiate irrelevant variables, and they abstract away the numbering of interchangeable objects.

Sec. 5 discusses the data structures and algorithms necessary to make generic M-H efficient. The goal here is to avoid having the time required to compute the acceptance probability and update the MCMC state grow with the number of hypothesized objects or the number of instantiated variables. We use data structures that represent a proposed world as a set of differences with respect to the current world (Sec. 5.1). More interestingly, we can determine which factors in the acceptance probability need to be recomputed by maintaining a context-specific Bayes net graph over the instantiated variables (Sec. 5.2). Sec. 6 presents experimental results on the citation matching task, showing that our generic M-H system supports the same proposal distribution that was used in a hand-coded implementation, and has a running time of the same order of magnitude.

## 2 BACKGROUND

### 2.1 CITATION MATCHING

Pasula et al. [2003] employ an M-H algorithm to cluster citations into groups that refer to the same publication. They use a generative model to define a distribution over worlds containing publications, researchers, and citations. This model includes prior distributions for researcher names and publication titles. The authors of each publication are chosen uniformly from the set of researchers, and the publication cited by each citation is chosen uniformly from the set of publications. The text of a citation is generated from the author names and title of the publication it cites, through an observation model that allows abbreviations, errors, and various styles of formating. Given the text of some observed citations, the goal is to infer which citations co-refer.

The proposal distribution of [Pasula *et al.*, 2003] is based on moves that split and merge clusters of co-referring citations. In a preprocessing step, the citations are grouped into *canopies*: overlapping groups of citations that are similar enough (according to a rough heuristic) that they have a non-negligible chance of co-referring [McCallum *et al.*, 2000]. For each move, the proposal distribution randomly chooses a canopy and two citations $c_1, c_2$ in that canopy. If $c_1$ and $c_2$ refer to the same publication $p_1$, then a new publication $p_2$ is added and the citations of $p_1$ are split randomly among $p_1$ and $p_2$. If $c_1$ and $c_2$ refer to different publications, then the two publications may be merged. New titles and authors are proposed for the affected publications, based on the text of citations that now refer to them. Finally, any publications and researchers that are no longer connected to citations are removed from the MCMC state.

The resulting M-H algorithm recovers between 93% and 97% of the true co-referring clusters exactly. This accuracy is significantly better than that reported by the developers of Citeseer [Lawrence *et al.*, 1999], and is competitive with more recent discriminative methods [Wellner *et al.*, 2004].

### 2.2 BAYESIAN LOGIC

Bayesian logic (BLOG) [Milch *et al.*, 2005a] is a language for defining probability distributions over structures that can contain varying numbers of objects with varying re-

```
1  type Res; type Pub; type Cit;

2  guaranteed Cit Cit1, Cit2, Cit3, Cit4;

3  #Res ~ NumResearchersPrior;
4  random String Name(Res r) ~ NamePrior;

5  #Pub ~ NumPublicationsPrior;
6  random String Title(Pub p) ~ TitlePrior;
7  random NaturalNum NumAuthors(Pub p)
8      ~ NumAuthorsPrior;
9  random Res NthAuthor(Pub p, NaturalNum n)
10     if (n < NumAuthors(p))
11        then ~ Uniform(Res r);

12 random Pub PubCited(Cit c)
13     ~ Uniform(Pub p);
14 random String TitleText(Cit c)
15     ~ TitleObsModel(Title(PubCited(c)));
16 random String NthAuthorText(Cit c,
17                             NaturalNum n)
18     if (n < NumAuthors(PubCited(c)))
19        then ~ AuthorObsModel
20           (Name(NthAuthor(PubCited(c), n)));

21 random String AuthListSuffix(Cit c,
22                              NaturalNum n)
23     if (n < NumAuthors(PubCited(c)))
24        then ~ AuthJoinModel
25           (NthAuthorText(c, n),
26            AuthListSuffix(c, Succ(n)))
27     else = "";

28 random String Text(Cit c)
29     ~ FormatModel(TitleText(c),
30                   AuthListSuffix(c, 0));
```

Figure 1: BLOG model for citation matching.

lations among them. These "possible worlds" are represented formally as *model structures* of a *typed first-order logical language*. A typed first-order language includes a set of *types*, such as String, NaturalNum, Citation, and Publication, and a set of *function symbols*[1], such as Name, Title, and PubCited. A model structure specifies a set of objects for each type, and for each function symbol, it specifies a mapping from argument tuples to values.

A BLOG model specifies a generative process for constructing such possible worlds. Fig. 1 shows a simplified version of the BLOG model that we use for citation matching. The model begins by declaring object types (line 1) and introducing some *guaranteed* objects that exist in all possible worlds. The rest of the model consists of two kinds of statements. *Number statements*, appearing on lines 3 and 5, describe generative steps that add new objects to the world. *Dependency statements* describe steps that set the value of a function on a tuple of arguments. For instance, the statement on lines 16–20 specifies a distribution for the observed author name NthAuthorText$(c, n)$, conditioning on the true name Name(NthAuthor(PubCited$(c), n)$).

The objects that exist in a possible world include built-in objects (of types NaturalNum, String, etc.), guaranteed objects introduced in the BLOG model, and *non-guaranteed* objects generated by number statements. Specifically, if a number statement for type $\tau$ generates $N$ objects, these objects are consecutively numbered pairs $(\tau, 1), \ldots, (\tau, N)$.

---

[1]We treat predicates as function symbols with return type Boolean, and constant symbols as zero-ary function symbols.

A possible world is fully specified by the values of certain *basic random variables*. For a given BLOG model, the basic variables include a *number variable* for each number statement,[2] and a *function application variable* $V_f[o_1, \ldots, o_k]$ for each $k$-ary random function $f$ and each tuple of appropriately typed objects $o_1, \ldots, o_k$ that exist in any possible world. A BLOG model implicitly defines a Bayes net (BN) over its basic random variables. This BN may be infinite, and it is typically *contingent*: edges are active only in certain contexts [Milch *et al.*, 2005b]. For instance, TitleText(Cit1) depends on Title((Pub, 7)) only in the context where PubCited(Cit1) = (Pub, 7).

## 3 METROPOLIS-HASTINGS ARCHITECTURE

Our general-purpose M-H system is implemented in Java as part of the BLOG inference engine, which is available for download from http://www.cs.berkeley.edu/~milch/blog. Proposal distributions are written as Java classes that implement an interface called Proposer. More precisely, a class implementing the Proposer interface defines a family of proposal distributions: an object of this class defines a specific distribution $q(s'|s_n)$ once it has been initialized with a particular BLOG model, observed evidence, and list of queries. Useful Proposer classes may be specialized to a particular model and type of query: for instance, a proposer for citation matching might only support BLOG models identical to that in Fig. 1 (except for variations in the number of citations); evidence that consists of observed text for each citation; and queries for whether two citations refer to the same publication (*i.e.*, have the same PubCited value).

To begin the MCMC chain, the M-H engine calls a method on the Proposer object that generates an initial state $s_0$ consistent with the evidence. Then on each iteration, the engine calls the following Proposer method:

    double proposeNextState(MCMCState state);

The state object passed in is a copy of the current state $s_n$. The method changes the values of certain basic random variables in state so that it represents a proposed state $s'$ chosen from $q(s'|s_n)$. The proposer must ensure that $s'$ satisfies the evidence and is complete enough to answer the queries specified at initialization. The method returns the log proposal ratio $\ln(q(s_n|s')/q(s'|s_n))$.

---

[2]In models where objects generate other objects, there is a number variable for each application of a number statement to a tuple of generating objects (see [Milch *et al.*, 2005a]).

The general-purpose M-H engine then computes the probability ratio $p(s')/p(s_n)$, and uses this along with the log proposal ratio to compute the acceptance probability given in Eq. 1. If it accepts the proposal, it sets $s_{n+1}$ equal to the `state` object that was modified by the proposer; otherwise, it sets $s_{n+1}$ equal to a saved copy of $s_n$.

## 4 MCMC STATES

MCMC states serve as the interface between the general-purpose and application-specific parts of the generic M-H system. The obvious way to apply MCMC to a BLOG model is to let the MCMC states be possible worlds. However, the proposal distribution of [Pasula *et al.*, 2003] does not propose complete possible worlds. A full possible world typically contains many publications that are not cited, that is, are not the value of PubCited($c$) for any citation $c$. The world must specify the values of the Title and NthAuthor functions on all the publications that exist. But the proposal distribution discussed in Sec. 2.1 never proposes titles or authors for uncited publications.

The contingent BN for the citation matching model makes it clear that attributes of uncited publications are irrelevant: in a world where a publication $p$ is uncited, the Title and NthAuthor variables defined on $p$ are not active ancestors of query or evidence variables. Proposing values for these variables would be a waste of time. In fact, in some BLOG models — such as a model for aircraft tracking with variables State($a, t$) for every aircraft $a$ and natural number $t$ — each possible world assigns non-null values to infinitely many variables. In such models, proposing and storing full possible worlds would require infinite time and space.

### 4.1 EVENTS AS MCMC STATES

Our generic MCMC architecture circumvents these difficulties by allowing proposal distributions to use *partial* descriptions of possible worlds. For instance, the proposer for citation matching specifies the values of the PubCited function on all citations, and specifies attributes for the cited publications and their authors. Such a partial specification can be thought of as an event: a set of full possible worlds that satisfy the specification.

Thus, our system runs a Markov chain over a set $\Sigma$ of *events*, which are subsets of the outcome space $\Omega$. The following theorem gives conditions under which a Markov chain over $\Sigma$ will yield correct answers to queries.

**Theorem 1.** *Let $p$ be a probability distribution over a set $\Omega$, $E$ and $Q$ be subsets of $\Omega$, and $\Sigma$ be a set of subsets of $\Omega$. Suppose $s_1, s_2, \ldots, s_N$ are samples from an ergodic Markov chain over $\Sigma$ with stationary distribution proportional to $p(s)$. If:*

1. *$\Sigma$ is a partition of $E$; and*

2. *for each $s \in \Sigma$, either $s \subseteq Q$ or $s \cap Q = \varnothing$,*

*then $\frac{1}{N} \sum_{n=1}^{N} \mathbf{1}(s_n \subseteq Q)$ converges to $p(Q|E)$.*

*Proof.* Let $\pi$ be the stationary distribution of the Markov chain over $\Sigma$, and let $\widetilde{Q}$ be the set of states $\{s \in \Sigma : s \subseteq Q\}$. Then by standard results about ergodic Markov chains, $\frac{1}{N} \sum_{n=1}^{N} \mathbf{1}(s_n \subseteq Q)$ converges to $\pi(\widetilde{Q})$ as $N \to \infty$. So it suffices to show that $\pi(\widetilde{Q}) = p(Q|E)$. By definition, $\pi(\widetilde{Q}) = \sum_{s \in \widetilde{Q}} \pi(s)$. Now since $\pi(s)$ is proportional to $p(s)$:

$$\pi(\widetilde{Q}) = \frac{\sum_{s \in \widetilde{Q}} p(s)}{\sum_{s \in \Sigma} p(s)} \quad (2)$$

By the assumption that $\Sigma$ is a partition of $E$, we know $\sum_{s \in \Sigma} p(s) = p(E)$. We now argue that the set of events $\widetilde{Q}$ is a partition of $Q \cap E$. To see this, consider any $\omega \in Q \cap E$. Because $\Sigma$ is a partition of $E$, there is exactly one set $s \in \Sigma$ such that $\omega \in s$. Given that $\omega \in s \cap Q$, it follows by assumption 2 that $s \subseteq Q$. Therefore $s \in \widetilde{Q}$. Thus, since $\widetilde{Q} \subseteq \Sigma$, there is exactly one $s \in \widetilde{Q}$ containing $\omega$. So $\widetilde{Q}$ is a partition of $Q \cap E$ and $\sum_{s \in \widetilde{Q}} p(s) = p(Q \cap E)$. Plugging into Eq. 2, we find that $\pi(\widetilde{Q}) = \frac{p(Q \cap E)}{p(E)} = p(Q|E)$. □

The next section discusses a way to choose the event set $\Sigma$.

### 4.2 PARTIAL INSTANTIATIONS

The most straightforward events to use as MCMC states are those corresponding to partial instantiations of the basic random variables. To satisfy Thm. 1, these partial instantiations must instantiate the evidence variables to their observed values, instantiate the query variables, and define a partition of the worlds consistent with the evidence.

Furthermore, to compute the acceptance probability given in Eq. 1, the system must be able to compute the ratio $p(s')/p(s_n)$ for events $s_n, s' \in \Sigma$. In general, it is not easy to compute the probability of a partial instantiation: for instance, if the instantiation just includes the evidence variables, then computing its probability involves summing out all the hidden variables. In some cases it is possible to sum out uninstantiated variables analytically, but our generic MCMC system currently cannot do so.

Instead, we limit ourselves to partial instantiations whose probabilities are given by simple product expressions. These are the *self-supporting* instantiations: those that include all the active parents of the variables they instantiate. To say this formally, we need a bit more background on contingent BNs (see [Milch *et al.*, 2005b] for details). In a contingent BN, the conditional probability distribution (CPD) for a variable $V$ is given by a tree where each internal node is labeled with a parent variable $U$, edges out of a node are labeled with values of $U$, and each leaf is labeled with a probability distribution over $V$. A particular parent variable may occur on some paths through the tree and not on others: for instance, in the tree for TitleText(Cit1), the root is labeled with PubCited(Cit1), and the variable Title((Pub, 7)) occurs only in the subtree where PubCited(Cit1) = (Pub, 7). An instantiation $\sigma$ *supports* $V$ if it is complete enough so that only one path through the tree is consistent with $\sigma$. This path leads to a leaf with some distribution over $V$; we write $p_V(v|\sigma)$ for the probability of the value $v$ under this distribution.

An instantiation is *self-supporting* if it supports every variable that it instantiates. By the semantics of a contingent BN, if $\sigma$ is a finite, self-supporting instantiation, then:

$$p(\sigma) = \prod_{V \in \text{vars}(\sigma)} p_V(\sigma(V)|\sigma) \quad (3)$$

where $\sigma(V)$ is the value that $\sigma$ assigns to $V$. Thus, if we use self-supporting partial instantiations as our MCMC states, we can compute $p(s')/p(s_n)$ with no summations.

To satisfy the conditions of Thm. 1, we need to use self-supporting instantiations that form a partition of $E$. In particular, we need to ensure that these instantiations are mutually exclusive: if some of them define overlapping events, then worlds occurring in several events will be over-counted. The following result ensures that we can avoid overlaps by using "minimal" instantiations.

**Definition 1.** *Let* **V** *be a set of random variables, and* $\sigma$ *be a self-supporting instantiation that instantiates* **V**. *Then* $\sigma$ *is* minimal beyond **V** *if no sub-instantiation of* $\sigma$ *that instantiates* **V** *is self-supporting.*

**Proposition 2.** *Let* **V** *be a set of random variables in a contingent BN. The self-supporting instantiations that are minimal beyond* **V** *are mutually contradictory.*

*Proof.* Assume for contradiction that two distinct self-supporting instantiations $\sigma$ and $\tau$ that are minimal beyond **V** are both satisfied by some world $\omega$. By definition, neither $\sigma$ nor $\tau$ can be a sub-instantiation of the other. Therefore $\sigma$ instantiates a variable, call it $X^*$, that $\tau$ does not instantiate. Consider a graph over $\text{vars}(\sigma)$ where there is an edge from $X$ to $Y$ if the path through $Y$'s CPD tree that is consistent with $\omega$ contains a node labeled with $X$. Since $\sigma$ is minimal beyond **V**, there must be a directed path in this graph from $X^*$ to **V**; otherwise the sub-instantiation obtained by removing $X^*$ and all its descendents would still instantiate **V** and be self-supporting. But since $\tau$ is also consistent with $\omega$, $\tau$ must instantiate all the variables along this directed path in order to be self-supporting. This contradicts the assumption that $\tau$ does not instantiate $X^*$. □

We have now identified a set of partial instantiations that satisfy the conditions of Thm. 1 and have probabilities that are easy to compute. If the evidence variables are $\mathbf{V}_E$ and the query variables are $\mathbf{V}_Q$, we use the set of self-supporting instantiations that assign the observed values to $\mathbf{V}_E$ and are miminal beyond $\mathbf{V}_E \cup \mathbf{V}_Q$.

### 4.3 OBJECT IDENTIFIERS

#### 4.3.1 Proposers and interchangeable objects

Recall that in a BLOG model, the objects that satisfy a given number statement are numbered. For instance, in worlds where there are 10 publications, the publication objects are (Pub, 1), ..., (Pub, 10). The BLOG model in Fig. 1 specifies that the value of each PubCited variable is chosen uniformly from these publication objects.

However, our description of the Pasula *et al.* proposal distribution in Sec. 2.1 does not say how it chooses publication objects to serve as the values for PubCited variables.[3] For instance, consider an MCMC state where #Pub = 1000, but only 200 distinct publication objects currently serve as values for PubCited variables. When the proposer performs a split move — taking, say (Pub, 7), and splitting off some of its citations to join a new publication — which of the 800 previously uncited publications is used as the PubCited value for these citations?

One answer that seems reasonable is to choose the lowest-numbered uncited publication. However, in order to have non-zero acceptance probabilities, all our MCMC moves must be reversible: the reverse proposal probability $q(s_n|s')$ must be positive. Under the policy just described, a move that merges, say, (Pub, 7) into (Pub, 3) is not reversible when (Pub, 1) happens to be uncited. Any split move in the resulting state would assign citations to (Pub, 1), not (Pub, 7).

Such reversibility problems can be avoided by choosing the new PubCited value *randomly* from the publications that are uncited in the current partial instantiation. But it seems that it should not be necessary to spend time invoking the pseudo-random number generator to choose a publication, since all the uncited publications are interchangeable. Furthermore, if the proposer samples publications randomly, then it must include these sampling probabilities in the forward and backward proposal probabilities; this increases the amount of bookkeeping involved in writing a proposal distribution.

#### 4.3.2 Abstract partial instantiations

Our general MCMC system includes an additional layer of abstraction that makes it easier to write proposal distributions involving interchangeable objects. The idea is to specify MCMC states using *abstract* partial instantiations, in which unnumbered *object identifiers* can be used as both arguments and values for basic random variables. For instance, an abstract partial instantiation using the identifier Pub@A3F could say: PubCited(Cit1) = Pub@A3F, Title(Pub@A3F) = "foo". We will refer to guaranteed and non-guaranteed objects that exist in possible worlds as *concrete objects*, to distinguish them from object identifiers.

---
[3]The original formulation of this proposer in [Pasula *et al.*, 2003] does not include PubCited variables; the MCMC states just specify a partition of the citations into co-referring groups.

**Definition 2.** *An* abstract function application variable *has the form $A_f[o_1, \ldots, o_k]$ where $f$ is a $k$-ary function symbol and $o_1, \ldots, o_k$ are concrete objects or object identifiers. An* abstract partial instantiation $\sigma$ *consists of a set of number variables[4] and abstract function application variables, denoted* vars$(\sigma)$, *and a function that maps each element of* vars$(\sigma)$ *to a concrete object or object identifier. For each type, an abstract partial instantiation uses either object identifiers or concrete objects to represent the non-guaranteed objects, not both.*

Semantically, object identifiers can be thought of as existentially quantified logical variables. The abstract partial instantiation used as an example above is equivalent to $\exists x((\mathsf{PubCited}(\mathsf{Cit1}) = x) \wedge (\mathsf{Title}(x) = \text{``foo''}))$. When an abstract instantiation uses several object identifiers, they are also asserted to be distinct.

**Definition 3.** *A* partial instantiation $\gamma$ *is a* concrete version *of an abstract partial instantiation $\sigma$ if there is a one-to-one function $h$ from object identifiers used in $\sigma$ to concrete objects that exist in some world consistent with $\gamma$, such that $\sigma$ instantiates $A_f[o_1, \ldots, o_k]$ if and only if $\gamma$ instantiates $V_f[h(o_1), \ldots, h(o_k)]$, and $h(\sigma(A_f[o_1, \ldots, o_k])) = \gamma(V_f[h(o_1), \ldots, h(o_k)])$. A* world satisfies *an abstract partial instantiation $\sigma$ if and only if it satisfies some concrete version of $\sigma$.*

For instance, the abstract partial instantiation:

#Pub = 3, PubCited(Cit1) = Pub@A3F, Tit(Pub@A3F) = "foo"

has three concrete versions:

#Pub = 3, PubCited(Cit1) = (Pub, 1), Tit((Pub, 1)) = "foo"
#Pub = 3, PubCited(Cit1) = (Pub, 2), Tit((Pub, 2)) = "foo"
#Pub = 3, PubCited(Cit1) = (Pub, 3), Tit((Pub, 3)) = "foo"

#### 4.3.3 Probabilities of abstract instantiations

Each abstract instantiation corresponds to an event, namely the set of possible worlds that satisfy it (note that two instantiations using different object identifiers may represent the same event). But how do we compute the probability of this event?

**Lemma 3.** *Let $\sigma$ be an abstract partial instantiation for a BLOG model. Then all concrete versions of $\sigma$ have the same probability, which we will call $p_c(\sigma)$. Also, if any concrete version of $\sigma$ is self-supporting, then they all are.*

The proof of this lemma relies on the stipulation in Def. 2 that an abstract instantiation cannot use both concrete objects and object identifiers for the same type. If such mixing were allowed, then some concrete versions might have

---
[4]In models where objects generate other objects, number variables can also be abstract.

---

different probabilities than others, depending on whether certain object identifiers were mapped to concrete objects used elsewhere in the instantiation.

If an abstract instantiation $\sigma$ contains an instantiated number variable asserting that there are $n$ objects of a given type, and $\sigma$ uses $m$ object identifiers of that type, then there are $_nP_m \triangleq \frac{n!}{(n-m)!}$ distinct functions that could play the role of $h$ in Def. 3. However, this observation does not lead to a general formula for the probability of $\sigma$. The problem is that the concrete versions produced by different $h$ functions may correspond to overlapping events. For instance, two concrete versions of the abstract instantiation (Title(Pub@A3F) = "foo") are (Title((Pub, 3)) = "foo") and (Title((Pub, 7)) = "foo"). There are many worlds that satisfy both these concrete instantiations. Two $h$ functions may even yield exactly the same concrete instantiation. For example, suppose $\sigma =$ (Title(Pub@A3F) = "foo", Title(Pub@B46) = "foo"). Here an $h$ function that maps Pub@A3F to (Pub, 1) and Pub@B46 to (Pub, 2) yields the same concrete instantiation as one that does the opposite, since $\sigma$ makes the same assertion about Pub@A3F and Pub@B46.

The difficulty in this last example is that $\sigma$ has a non-trivial automorphism: interchanging Pub@A3F and Pub@B46 yields $\sigma$ itself. In general, the number of distinct concrete versions of an abstract instantiation with $a$ automorphisms is $\frac{1}{a}\left(_nP_m\right)$. But if the instantiation specifies relations among the non-guaranteed objects — for instance, publications citing one another — then counting automorphisms becomes difficult. Indeed, counting the number of automorphisms of an undirected graph is polynomially equivalent to determining whether two graphs are isomorphic [Mathon, 1979], a problem for which no polynomial-time algorithm is known. This issue of automorphisms does not just arise because we are trying to use abstract partial instantiations as MCMC states: if we required the proposer to choose non-guaranteed objects randomly, its proposal probability calculations would also need to determine how many different choices would yield the same proposal.

#### 4.3.4 A useful special case

Fortunately, for many models of practical interest, there is a simple way to avoid this issue. The abstract partial instantiations that we use for citation matching only make assertions about cited publications. That is, if $\sigma$ uses an object identifier such as Pub@A3F, then $\sigma$ asserts PubCited($c$) = Pub@A3F for some citation $c$. If we apply two $h$ functions that yield different concrete values for Pub@A3F, say (Pub, 1) and (Pub, 2), then the resulting concrete versions define disjoint events: one asserts PubCited($c$) = (Pub, 1) and the other asserts PubCited($c$) = (Pub, 2). In general:

**Definition 4.** *An object identifier $i$ is* grounded *in an ab-*

stract partial instantiation $\sigma$ if there is a logical ground term $t_i$ such that every mapping function $h$ (as in Def. 3) yields a concrete instantiation where $h(i)$ is the value of $t_i$.

**Proposition 4.** *Suppose $\sigma$ is an abstract partial instantiation whose concrete versions are self-supporting instantiations having probability $p_c(\sigma)$. Let $T$ be the set of types for which $\sigma$ uses identifiers, and assume that for every type $\tau \in T$, $\sigma$ instantiates a number variable asserting that there are $n_\tau$ non-guaranteed objects of type $\tau$. If every identifier used in $\sigma$ is grounded, then:*

$$p(\sigma) = p_c(\sigma) \prod_{\tau \in T} n_\tau P_{m_\tau} \quad (4)$$

*where $m_\tau$ is the number of identifiers of type $\tau$ used in $\sigma$.*[5]

*Proof.* Given our discussion above, it suffices to show that the mapping functions $h$ all yield disjoint events when applied to $\sigma$. Consider any two distinct mapping functions $h_1$ and $h_2$, and let $\gamma_1$ and $\gamma_2$ be the corresponding concrete versions of $\sigma$. Let $i$ be any identifier used in $\sigma$ such that $h_1(i) \neq h_2(i)$. Because $i$ is grounded in $\sigma$, Def. 4 implies that there is a logical ground term $t_i$ such that $t_i$ evaluates to $h_1(i)$ in every world satisfying $\gamma_1$, and $t_i$ evaluates to $h_2(i)$ in every world satisfying $\gamma_2$. Since $h_1(i) \neq h_2(i)$, this implies $\gamma_1$ and $\gamma_2$ are disjoint. □

Logical ground terms include not just expressions such as PubCited(Cit1), but also nested expressions such as NthAuthor(PubCited(Cit1), 1). The requirement that object identifiers be grounded is not burdensome in scenarios — such as citation matching — where the relevant objects are those connected to guaranteed objects by some chains of function applications. In BLOG models that involve weighted sampling or aggregation, non-guaranteed objects that do not serve as function values may become relevant. In such cases, the proposer would need to represent such objects concretely.

In cases where Prop. 4 applies, we can compute the probability of an abstract instantiation by just computing the probability of one of its concrete versions and then multiplying in an adjustment factor that is a product of factorials. In fact, the ratio of these adjustment factors in the acceptance probability is the same as the ratio of adjustments to the backward and forward proposal probabilities that would emerge if we required the proposal distribution to choose a distinct non-guaranteed object for each identifier randomly. But we have avoided the need to actually do this random sampling, and shifted this computation from the application-specific proposal distribution to general-purpose code.

---

[5]This result can be extended to cases where objects generate objects; then the product is not over types, but over applications of number statements to tuples of generating objects. Abstract instantiations must be extended to specify the generating objects for each object identifier.

## 5 PERFORMING M-H STEPS EFFICIENTLY

Our overall goal is to compute the probability ratio and update the MCMC state in time that does not grow with the number of existing objects or the number of instantiated variables. This is not always possible, but application-specific implementations exploit various forms of structure to do these computations in constant time. We are able to exploit some of the same structure in our generic system.

### 5.1 DIFFERENCE DATA STRUCTURES

We said in Sec. 3 that the M-H engine saves a copy of the current state $s_n$ before passing a modifiable copy to the proposer. But making a full copy of $s_n$ would take time linear in the number of instantiated variables. Thus, our implementation does something more subtle. The `state` object passed to `proposeNextState` is actually a difference structure or "patch" built on top of the current state $s_n$. This difference structure contains a hash table that maps changed or newly instantiated basic variables to their new values, as well as a list of newly uninstantiated variables. The proposer actually just changes this patch; the underlying copy of $s_n$ is left unchanged. However, the difference structure supports all the same access methods as an ordinary `MCMCState` data structure: if a client asks for the value of a variable that has not been changed, the request is just passed through to the original state.

If the proposal is rejected, the patch is simply discarded, and $s_{n+1}$ is set equal to $s_n$. If the proposal is accepted, then $s_{n+1}$ is obtained by applying the patch to $s_n$: that is, changing the underlying state so it reflects the changes made in the patch. This operation takes time linear in the number of changed variables. The patch is then cleared, leaving it free to accept modifications from the next call to `proposeNextState`.

### 5.2 COMPUTING THE ACCEPTANCE PROBABILITY

Besides maintaining the MCMC state, the main task for our general-purpose code is to compute the acceptance probability. The proposal distribution provides $q(s_n|s')/q(s'|s_n)$, so we must compute the probability ratio $p(s')/p(s_n)$. If $s_n$ and $s'$ are represented as self-supporting partial instantiations $\sigma_n$ and $\sigma'$, Eq. 3 tells us that this ratio is:

$$\frac{p(\sigma')}{p(\sigma_n)} = \frac{\prod_{V \in \text{vars}(\sigma')} p_V(\sigma'(V)|\sigma')}{\prod_{V \in \text{vars}(\sigma_n)} p_V(\sigma_n(V)|\sigma_n)} \quad (5)$$

Computing this ratio naively would require time proportional to the number of instantiated variables in $\sigma'$ and $\sigma_n$. But fortunately, many of the factors in the numerator and denominator may cancel.

**Definition 5.** *If a partial instantiation $\sigma$ supports a variable $V$, then the active parents of $V$ in $\sigma$ are those variables that occur as labels on nodes in $V$'s CPD tree on the unique path that is consistent with $\sigma$.*

**Proposition 5.** *Suppose two partial instantiations $\sigma$ and $\sigma'$ agree on a variable $V$ and on all the variables that are active parents of $V$ in $\sigma$. Then $p_V(\sigma'(V)|\sigma') = p_V(\sigma(V)|\sigma)$. Also, $V$ has the same active parents in $\sigma'$ as in $\sigma$.*

Thus, we only need to compute the factors for variables that are newly instantiated, uninstantiated, or changed in $\sigma'$, or whose active parents have changed values. Because we are explicitly representing the differences between $\sigma'$ and $\sigma_n$ (see Sec. 5.1), we can identify the changed variables efficiently. However, it is not so easy to identify variables whose active parents have changed. We can enumerate $V$'s active parents in $\sigma_n$ by walking through $V$'s CPD tree. But if only a few variables have changed, we don't want to iterate over all variables, seeing which ones happen to have a changed variable as an active parent.

To avoid this iteration, we maintain a graph over the instantiated variables in $\sigma_n$; this graph contains those edges from the BLOG model's contingent BN that are active in $\sigma_n$. Each variable has pointers to its children, that is, the variables of which it is an active parent in $\sigma_n$. Given this data structure, we can efficiently enumerate the children of all variables that are changed in $\sigma'$. The graph is constructed on the initial state, and then updated after each accepted proposal to reflect newly active or inactive parent relationships in $\sigma'$. Conveniently, by Prop. 5, we only need to update a variable's active parent set if one of its active parents in $\sigma_n$ has changed – and we are enumerating these variables anyway to recompute their probability factors.

If we use abstract partial instantiations, the probability ratio includes the factorial adjustment factors given in Eq. 4. Again, computing these factorials naively would take time linear in the magnitudes of the number variables. But if the proposal makes small changes to the values of number variables and the number of used identifiers, then most of the factors inside the factorials cancel out.

The calculation techniques presented here do have some limitations. One is that a variable's child set may grow linearly with the number of objects. In the citation matching model, where the probability that a PubCited variable takes on any particular value in a world with $N$ publications is $1/N$, the #Pub variable is always an active parent of all PubCited variables. So the time required to compute the acceptance probability for a proposal that changes the number of publications grows linearly with the number of citations. This slowdown could be avoided by recognizing that every PubCited variable makes the same contribution the probability ratio, so we can compute this contribution once and raise it to the power of the number of citations. However, our current implementation does not detect when this can be done. Conversely, a variable's active parent set may grow linearly with the number of hypothesized objects: this happens in cases of weighted sampling or aggregation. Finally, our approach does not allow the system to detect cancellations between the $p(s')$ and $q(s'|s_n)$ factors, such as occur in Gibbs sampling [Gelman, 1992].

## 6 EXPERIMENTS

We have developed a BLOG model and proposal distribution for the citation matching domain. Earlier work on applying M-H to this task [Pasula *et al.*, 2003] used an implementation hand-coded in Lisp. Unfortunately, we do not know all the details of the model and proposal distribution used in that implementation, nor do we have data on its running time. However, we do have all these details for an application-specific Java system that we implemented in the summer of 2003. Our BLOG implementation almost exactly reproduces the model and proposal distribution used in this hand-coded Java system, which thus serves as our reference for speed and accuracy comparisons.

The BLOG model we use is an elaboration of the one shown in Fig. 1. The prior distributions for author names and titles are n-gram models learned from the author and title fields of a large BibTeX file; some parameters of the citation formating model are estimated from a set of hand-segmented citations. Other parameters, such as typo probabilities, are set by hand; the exact values of these parameters have little influence on the accuracy results. The proposal distribution uses split-merge moves of the kind described in [Jain and Neal, 2004].

Table 1 shows results on four sets of about 300–500 unparsed citations that were collected by [Lawrence *et al.*, 1999]. The files are annotated with the true clustering of citations into co-referring groups; the accuracy metric is the fraction of true clusters recovered exactly. All of the M-H implementations achieve better accuracy than the [Lawrence *et al.*, 1999] technique, with the [Pasula *et al.*, 2003] implementation doing best by a significant margin. The [Pasula *et al.*, 2003] implementation outperforms the others because it uses more sophisticated prior distributions for author names and citation formats, and more finely tuned heuristics for proposing parses of citations. There is little difference in accuracy between the hand-coded Java implementation and the general-purpose BLOG engine; this is to be expected, since they implement approximately the same model and proposal distribution.

The timing results in Table 1 reflect the time required to initialize the system and run MCMC for 10,000 samples. Both systems display significant variation in run time across data sets; this reflects differences in the average number of citations affected by split-merge moves (the data sets have different ratios of citations to publications) and differences in the fraction of proposals that are accepted. However,

|  |  | Face<br>349 citations | Reinforcement<br>406 citations | Reasoning<br>514 citations | Constraint<br>295 citations |
|---|---|---|---|---|---|
| Phrase matching | accuracy | 94% | 79% | 86% | 89% |
| M-H: Pasula *et al.* | accuracy (avg) | 97% | 94% | 96% | 93% |
| M-H: Java | accuracy (avg) | 95.1 (±1.3)% | 81.8 (±2.5)% | 88.6 (±1.0)% | 91.7 (±1.4)% |
|  | accuracy (final) | 96.0 (±0.9)% | 86.1 (±3.6)% | 89.4 (±1.8)% | 91.8 (±2.1)% |
|  | time | 14.3 (±0.1) s | 19.4 (±0.2) s | 19.0 (±0.3) s | 12.1 (±0.1) s |
| M-H: BLOG | accuracy (avg) | 95.6 (±0.8)% | 78.0 (±2.3)% | 88.7 (±0.8)% | 90.7 (±0.9)% |
|  | accuracy (final) | 96.3 (±0.8)% | 82.3 (±2.1)% | 90.8 (±1.0) % | 91.9 (±0.9)% |
|  | time | 69.7 (±2.3) s | 99.0 (±4.1) s | 99.4 (±4.1) s | 59.9 (±0.7) s |

Table 1: Citation matching results for the phrase matching algorithm of [Lawrence *et al.*, 1999], the hand-coded M-H implementation used by [Pasula *et al.*, 2003], a simpler M-H implementation hand-coded in Java, and the BLOG inference engine. For the M-H algorithms, accuracy may be averaged over all 10,000 samples or computed on the final MCMC state; times are measured for a run that computes the accuracy only on the final state. For the last two systems, we give 95% confidence intervals based on 10 independent runs.

the BLOG engine consistently takes 5 times as long as the hand-coded Java implementation.

There are three main reasons for this difference. First, in the hand-coded implementation, an MCMC state is represented as a collection of Java objects of application-specific classes such as `Publication` and `Citation`. The current values of functions such as `Title` and `PubCited` are stored in fields on those objects. By contrast, the general BLOG engine does not include specialized Java classes for the citation domain; it uses a hash table that maps functions and argument tuples to values. Thus, accessing and updating the state is considerably slower in the BLOG implementation. Second, the hand-coded implementation includes special code for determining which variables are affected by moves that are proposed by its specific proposal distribution. In order to support arbitrary proposals, the BLOG engine must look at the list of variables changed by the proposal, find their children in the current BN graph, and (if the proposal is accepted) update the BN graph to reflect dependencies that are active in the new world. Finally, computing the probabilities of variables given their parents is slower in the BLOG implementation. The BLOG engine must interpret if–then clauses that occur in the BLOG model (e.g., lines 10 and 23 of Fig. 1) and explicitly store values in the MCMC state for intermediate variables, such as `AuthListSuffix` in Fig. 1. In the hand-coded implementation, if-statements and local variables can be written into the Java code, allowing faster execution.

As a result, while the hand-coded implementation does 10,000 samples in 12–20 seconds, the BLOG engine takes 60–100 seconds. However, our notes from summer of 2003 indicate that with the computers and Java runtime environment we had then, the hand-coded implementation ran in about 120 seconds. In other words, our computing infrastructure has improved enough that a general system runs faster than a hand-coded system did three years ago.

## 7 CONCLUSIONS

We have described a general MCMC inference system that just requires the user to provide a BLOG model and a proposal distribution. Our main contribution is a semantics for MCMC states that do not fully specify a possible world. By allowing partial world descriptions, we support proposal distributions that do not instantiate irrelevant variables or assign numbers to interchangeable objects. We also show how to use a context-specific Bayes net graph to determine efficiently what factors in the acceptance probability need to be computed for a given proposal.

Our current system still requires that the user implement a proposal distribution, which can be a significant undertaking. Other sampling-based approaches to approximate inference require less customization, and allow correspondingly less flexibility. For instance, the widely used BUGS system [Gilks *et al.*, 1994] allows users to run Gibbs sampling on a wide range of graphical models with no additional programming. However, Gibbs sampling falls short in scenarios where it is difficult to move between high-probability hypotheses by changing one variable at a time. In such cases, M-H algorithms can explore the posterior distribution more efficiently. There has been some recent work on adding generic M-H capabilities to BUGS using adaptive proposal distributions [Lunn *et al.*, 2005].

Another approach to automatic approximate inference is forward sampling: using the model's CPDs to sample variables given their parents. Milch *et al.* [2005b] use forward sampling in a general likelihood weighting algorithm for contingent BNs. Jaeger [2006] explores several variations on forward sampling for relational Bayesian networks, including a version where values are also propagated up from evidence nodes through deterministic dependencies. Angelopoulos and Cussens [2001; 2005], on the other hand, use forward sampling in a proposal distribution within an

M-H algorithm. Their models are represented as stochastic logic programs, which define distributions over Prolog proof trees; the proposal distribution resamples a sub-tree of the current proof tree. This algorithm has been successful on several applications. However, it seems that more data-driven proposal distributions are needed for applications such as citation matching, where forward sampling has a negligible probability of yielding author names and publication titles consistent with the observed citations.

Clearly there is more work to be done on general-purpose inference for relational probabilistic models. In the citation matching domain, we are extending our BLOG model and proposal distribution to simultaneously reconstruct the publications, researchers, and venues mentioned in a set of citations. We also plan to develop BLOG models and proposal distributions for other tasks, such as resolving coreference among names and pronouns in newswire articles. We hope that through these efforts, we will come to understand some common principles that underlie effective proposal distributions for various tasks. This understanding should lead toward the development of a library of (possibly adaptive) proposal distribution modules, which can be combined to yield effective proposal distributions for new tasks with little or no programming.

## Acknowledgements

This work was supported by DARPA IPTO under the CALO project (03–000219) and the Effective Bayesian Transfer Learning project (FA8750–05–2–0249). B. Milch was also supported by a Siebel Scholarship.